\documentclass[10pt,twocolumn,letterpaper]{article}

\usepackage{cvpr}
\usepackage{times}
\usepackage{epsfig}
\usepackage{graphicx}
\usepackage{amsmath}
\usepackage{amssymb}

\usepackage{booktabs}
\usepackage{tabulary}
\usepackage[dvipsnames]{xcolor}
\usepackage[caption=false]{subfig}
\usepackage[british,english,american]{babel}
\usepackage{soul}
\usepackage{xspace}\xspace
\usepackage{adjustbox}
\usepackage{array}

\usepackage{dblfloatfix}
\usepackage{transparent}

\usepackage{algorithm}
\usepackage{listings}

\usepackage{etoolbox}
\makeatletter
\AfterEndEnvironment{algorithm}{\let\@algcomment\relax}
\AtEndEnvironment{algorithm}{\kern2pt\hrule\relax\vskip3pt\@algcomment}
\let\@algcomment\relax
\newcommand\algcomment[1]{\def\@algcomment{\footnotesize#1}}
\renewcommand\fs@ruled{\def\@fs@cfont{\bfseries}\let\@fs@capt\floatc@ruled
  \def\@fs@pre{\hrule height.8pt depth0pt \kern2pt}%
  \def\@fs@post{}%
  \def\@fs@mid{\kern2pt\hrule\kern2pt}%
  \let\@fs@iftopcapt\iftrue}
\makeatother
\definecolor{citecolor}{HTML}{0071bc}
\usepackage[pagebackref=false,breaklinks=true,letterpaper=true,colorlinks,citecolor=citecolor,bookmarks=false]{hyperref}

\newcommand{\app}{\raise.17ex\hbox{$\scriptstyle\sim$}}

\newcommand{\apbbox}[1]{AP$^\text{bb}_\text{#1}$}
\newcommand{\apmask}[1]{AP$^\text{mk}_\text{#1}$}
\newcommand{\apkp}[1]{AP$^\text{kp}_\text{#1}$}
\newcommand{\apdp}[1]{AP$^\text{dp}_\text{#1}$}

\newcolumntype{x}[1]{>{\centering\arraybackslash}p{#1pt}}
\newcolumntype{y}[1]{>{\raggedright\arraybackslash}p{#1pt}}
\newcolumntype{z}[1]{>{\raggedleft\arraybackslash}p{#1pt}}
\newlength\savewidth\newcommand\shline{\noalign{\global\savewidth\arrayrulewidth
  \global\arrayrulewidth 1pt}\hline\noalign{\global\arrayrulewidth\savewidth}}
\newcommand{\tablestyle}[2]{\setlength{\tabcolsep}{#1}\renewcommand{\arraystretch}{#2}\centering\footnotesize}

\makeatletter\renewcommand\paragraph{\@startsection{paragraph}{4}{\z@}
  {.5em \@plus1ex \@minus.2ex}{-.5em}{\normalfont\normalsize\bfseries}}\makeatother

\def\x{\times}

\usepackage{textpos}  % for superposing a text box

\newcommand{\appdx}{appendix}

\cvprfinalcopy % *** Uncomment this line for the final submission

\def\cvprPaperID{***} % *** Enter the CVPR Paper ID here

\ifcvprfinal\fi
\begin{document}

%%%%%%%%% TITLE
\title{Momentum Contrast for Unsupervised Visual Representation Learning \vspace{-.5em}}

\author{
\vspace{.5em}
 Kaiming He \quad Haoqi Fan \quad Yuxin Wu \quad Saining Xie \quad Ross Girshick \vspace{.5em}\\
 Facebook AI Research (FAIR) \vspace{.3em}
}

\maketitle
%\thispagestyle{empty}

%%%%%%%%% ABSTRACT
\begin{abstract}
\vspace{-.5em}
We present Momentum Contrast (MoCo) for unsupervised visual representation learning. From a perspective on contrastive learning \cite{Hadsell2006} as dictionary look-up, we build a dynamic dictionary with a queue and a moving-averaged encoder. This enables building a large and consistent dictionary on-the-fly that facilitates contrastive unsupervised learning. MoCo provides competitive results under the common linear protocol on ImageNet classification. More importantly, the representations learned by MoCo transfer well to downstream tasks. MoCo can \textbf{outperform} its supervised pre-training counterpart in \textbf{7} detection/segmentation tasks on PASCAL VOC, COCO, and other datasets, sometimes surpassing it by large margins. This suggests that the gap between unsupervised and supervised representation learning has been largely closed in many vision tasks.

\vspace{-1em}
\end{abstract}

% hack: add link to code
\begin{textblock*}{.8\textwidth}[.5,0](0.5\textwidth, -.46\textwidth)
\centering
{\small Code: \url{https://github.com/facebookresearch/moco}}
\end{textblock*}

%%%%%%%%% BODY TEXT
\section{Introduction}

Unsupervised representation learning is highly successful in natural language processing, \eg, as shown by GPT \cite{Radford2018,Radford2019} and BERT \cite{Devlin2019}. But supervised pre-training is still dominant in computer vision, where unsupervised methods generally lag behind. The reason may stem from differences in their respective signal spaces. Language tasks have \mbox{discrete} signal spaces (words, sub-word units, \etc) for building tokenized \emph{dictionaries}, on which unsupervised learning can be based.
Computer vision, in contrast, further concerns dictionary building \cite{Sivic2003,Coates2011,Chatfield2011}, as the raw signal is in a continuous, high-dimensional space and is not structured for human communication (\eg, unlike words).

Several recent studies \cite{Wu2018a,Oord2018,Hjelm2019,Zhuang2019,Henaff2019,Tian2019,Bachman2019} present promising results on unsupervised visual representation learning using approaches related to the \emph{contrastive loss} \cite{Hadsell2006}. Though driven by various motivations, these methods can be thought of as building \mbox{\emph{dynamic dictionaries}}. The ``keys" (tokens) in the dictionary are sampled from data (\eg, images or patches) and are represented by an encoder network. \mbox{Unsupervised} learning trains encoders to perform dictionary look-up: an encoded ``query" should be similar to its matching key and dissimilar to others. Learning is formulated as minimizing a contrastive loss \cite{Hadsell2006}.

From this perspective, we hypothesize that it is desirable to build dictionaries that are: (i) large \emph{and} (ii) consistent as they evolve during training. Intuitively, a larger dictionary may better sample the underlying continuous, high-dimensional visual space, while the keys in the dictionary should be represented by the same or similar encoder so that their comparisons to the query are consistent. However, existing methods that use contrastive losses can be limited in one of these two aspects (discussed later in context).

%##################################################################################################
\begin{figure}[t]\centering
\vspace{1.em}
\includegraphics[width=.65\linewidth]{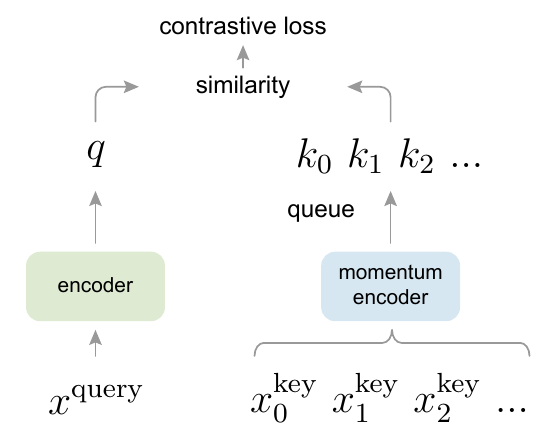}
\caption{Momentum Contrast (MoCo) trains a visual representation encoder by matching an encoded query $q$ to a dictionary of encoded keys using a contrastive loss. The dictionary keys $\{k_0, k_1, k_2, ...\}$ are defined on-the-fly by a set of data samples.
The dictionary is built as a queue, with the current mini-batch enqueued and the oldest mini-batch dequeued, decoupling it from the mini-batch size.
The keys are encoded by a slowly progressing encoder, driven by a momentum update with the query encoder.
This method enables a large and consistent dictionary for learning visual representations.
\label{fig:teaser}}
\vspace{-1em}
\end{figure}
%##################################################################################################

We present {\emph{Momentum Contrast}} (MoCo) as a way of building large and consistent dictionaries for unsupervised learning with a contrastive loss (Figure~\ref{fig:teaser}). We maintain the dictionary as a \emph{queue} of data samples: the encoded representations of the current mini-batch are enqueued, and the oldest are dequeued. The queue decouples the dictionary size from the mini-batch size, allowing it to be large. Moreover, as the dictionary keys come from the preceding several mini-batches, a \emph{slowly progressing} key encoder, implemented as a momentum-based moving average of the query encoder, is proposed to maintain consistency.

MoCo is a mechanism for building dynamic dictionaries for contrastive learning, and can be used with various pretext tasks. In this paper, we follow a simple instance discrimination task \cite{Wu2018a,Ye2019,Bachman2019}: a query matches a key if they are encoded views (\eg, different crops) of the same image.
Using this pretext task, MoCo shows competitive results under the common protocol of linear classification in the ImageNet dataset \cite{Deng2009}.

A main purpose of unsupervised learning is to pre-train representations (\ie, features) that can be transferred to downstream tasks by fine-tuning. We show that in 7 downstream tasks related to detection or segmentation, MoCo unsupervised pre-training can \emph{surpass} its ImageNet supervised counterpart, in some cases by nontrivial margins. In these experiments, we explore MoCo pre-trained on ImageNet or on a \emph{one-billion} \mbox{Instagram} image set, demonstrating that MoCo can work well in a more real-world, billion-image scale, and relatively uncurated scenario.
These results show that MoCo largely closes the gap between unsupervised and supervised representation learning in many computer vision tasks, and can serve as an alternative to ImageNet supervised pre-training in several applications.

\section{Related Work}

Unsupervised/self-supervised\footnote{Self-supervised learning is a form of unsupervised learning. Their distinction is informal in the existing literature. In this paper, we use the more classical term of ``unsupervised learning", in the sense of ``not supervised by \emph{human}-annotated labels".} learning methods generally involve two aspects: pretext tasks and loss functions.
The term ``pretext'' implies that the task being solved is not of genuine interest, but is solved only for the true purpose of learning a good data representation.
Loss functions can often be investigated independently of pretext tasks.
MoCo focuses on the loss function aspect. Next we discuss related studies with respect to these two aspects.

\paragraph{Loss functions.} A common way of defining a loss function is to measure the difference between a model's prediction and a \emph{fixed} target, such as reconstructing the input pixels (\eg, auto-encoders) by L1 or L2 losses, or classifying the input into pre-defined categories (\eg, eight positions \cite{Doersch2015}, color bins \cite{Zhang2016}) by cross-entropy or margin-based losses.
Other alternatives, as described next, are also possible.

Contrastive losses \cite{Hadsell2006} measure the similarities of sample pairs in a representation space. Instead of matching an input to a fixed target, in contrastive loss formulations the target can \emph{vary} on-the-fly during training and can be defined in terms of the data representation computed by a network \cite{Hadsell2006}.
Contrastive learning is at the core of several recent works on unsupervised learning \cite{Wu2018a,Oord2018,Hjelm2019,Zhuang2019,Henaff2019,Tian2019,Bachman2019}, which we elaborate on later in context (Sec.~\ref{sec:contrast}).

Adversarial losses \cite{Goodfellow2014} measure the difference between probability distributions. 
It is a widely successful technique for unsupervised data generation. Adversarial methods for representation learning are explored in \cite{Donahue2017,Donahue2019}. There are relations (see \cite{Goodfellow2014}) between generative adversarial networks and noise-contrastive estimation (NCE) \cite{Gutmann2010}.

\paragraph{Pretext tasks.} 
A wide range of pretext tasks have been proposed. Examples include recovering the input under some corruption, \eg, denoising auto-encoders \cite{Vincent2008}, context auto-encoders \cite{Pathak2016}, or cross-channel auto-encoders (colorization) \cite{Zhang2016,Zhang2017}.
Some pretext tasks form pseudo-labels by, \eg, transformations of a single (``exemplar'') image \cite{Dosovitskiy2014}, patch orderings \cite{Doersch2015,Noroozi2016}, tracking \cite{Wang2015a} or segmenting objects \cite{Pathak2017} in videos, or clustering features \cite{Caron2018,Caron2019}.

\paragraph{Contrastive learning \vs pretext tasks.} 
Various pretext tasks can be based on some form of contrastive loss functions.
The instance discrimination method \cite{Wu2018a} is related to the exemplar-based task \cite{Dosovitskiy2014} and 
NCE \cite{Gutmann2010}.
The pretext task in contrastive predictive coding (CPC) \cite{Oord2018} is a form of context auto-encoding \cite{Pathak2016}, and in contrastive multiview coding (CMC) \cite{Tian2019} it is related to colorization \cite{Zhang2016}. 

\section{Method}

\subsection{Contrastive Learning as Dictionary Look-up}
\label{sec:contrast}

Contrastive learning \cite{Hadsell2006}, and its recent developments, can be thought of as training an encoder for a \emph{dictionary look-up} task, as described next.

Consider an encoded query $q$ and a set of encoded samples $\{k_0, k_1, k_2, ...\}$ that are the keys of a dictionary. Assume that there is a single key (denoted as $k_{+}$) in the dictionary that $q$ matches. A contrastive loss \cite{Hadsell2006} is a function whose value is low when $q$ is similar to its positive key $k_{+}$ and dissimilar to all other keys (considered negative keys for $q$). With similarity measured by dot product, a form of a contrastive loss function, called \mbox{InfoNCE} \cite{Oord2018}, is considered in this paper:
\begin{equation}
\small
\mathcal{L}_q = -\log \frac{\exp(q{\cdot}k_+ / \tau)}{\sum_{i=0}^{K}\exp(q{\cdot}k_i  / \tau)}
\label{eq:infonce}
\end{equation}
where $\tau$ is a temperature hyper-parameter per \cite{Wu2018a}. 
The sum is over one positive and $K$ negative samples.
Intuitively, this loss is the log loss of a $({K}{+}1)$-way softmax-based classifier that tries to classify $q$ as $k_{+}$. 
Contrastive loss functions can also be based on other forms \cite{Hadsell2006,Wang2015a,Wu2018a,Hjelm2019}, such as margin-based losses and variants of NCE losses.

%##################################################################################################
\begin{figure*}[t]\centering
\includegraphics[width=0.7\linewidth]{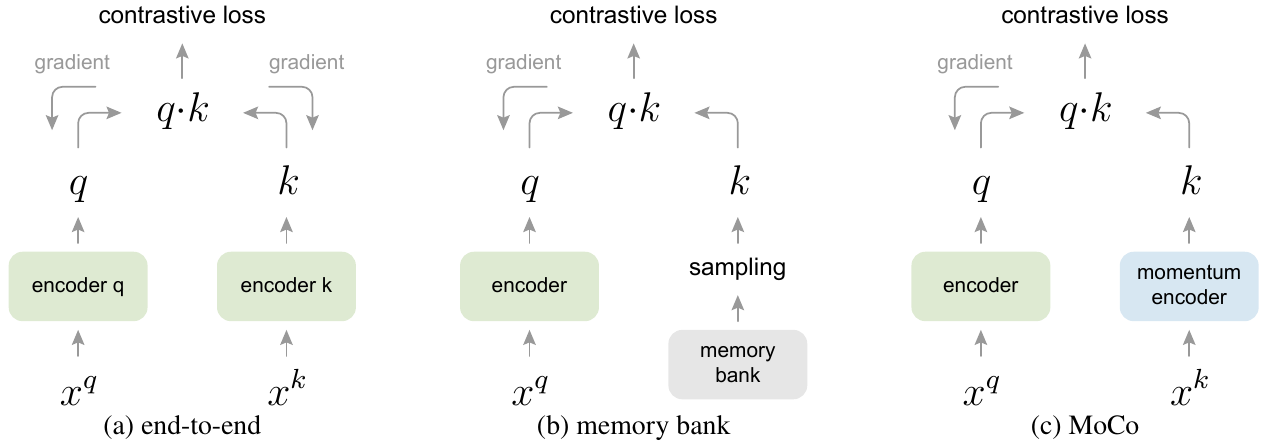}
\vspace{.1em}
\caption{\textbf{Conceptual comparison of three contrastive loss mechanisms} (empirical comparisons are in Figure~\ref{fig:results_mechanisms} and Table~\ref{tab:voc_mechanisms}).
Here we illustrate one pair of query and key.
The three mechanisms differ in how the keys are maintained and how the key encoder is updated.
\textbf{(a)}: The encoders for computing the query and key representations are updated \emph{\textbf{end-to-end}} by back-propagation (the two encoders can be different).
\textbf{(b)}: The key representations are sampled from a \emph{\textbf{memory bank}} \cite{Wu2018a}. \textbf{(c)}: \emph{\textbf{MoCo}} encodes the new keys on-the-fly by a momentum-updated encoder, and maintains a queue (not illustrated in this figure) of keys.
\label{fig:mechanisms}}
\vspace{-.5em}
\end{figure*}
%##################################################################################################

The contrastive loss serves as an unsupervised objective function for training the encoder networks that represent the queries and keys \cite{Hadsell2006}. In general, the query representation is $q=f_\textrm{q}(x^q)$ where $f_\textrm{q}$ is an encoder network and $x^q$ is a query sample (likewise, $k=f_\textrm{k}(x^k)$). 
Their instantiations depend on the specific pretext task.
The input $x^q$ and $x^k$ can be images \cite{Hadsell2006,Wu2018a,Ye2019}, patches \cite{Oord2018}, or context consisting a set of patches \cite{Oord2018}.
The networks $f_\textrm{q}$ and $f_\textrm{k}$ can be identical \cite{Hadsell2006,Wang2015a,Ye2019}, partially shared \cite{Oord2018,Hjelm2019,Bachman2019}, or different \cite{Tian2019}.

\subsection{Momentum Contrast}

From the above perspective, contrastive learning is a way of building a discrete dictionary on high-dimensional continuous inputs such as images. The dictionary is \emph{dynamic} in the sense that the keys are randomly sampled, and that the key encoder evolves during training.
Our hypothesis is that good features can be learned by a \emph{large} dictionary that covers a rich set of negative samples, while the encoder for the dictionary keys is kept as \emph{consistent} as possible despite its evolution. Based on this motivation, we present \mbox{Momentum} Contrast as described next.

\paragraph{Dictionary as a queue.} At the core of our approach is maintaining the dictionary as a \emph{queue} of data samples. This allows us to reuse the encoded keys from the immediate preceding mini-batches. The introduction of a queue  decouples the dictionary size from the mini-batch size. Our dictionary size can be much larger than a typical mini-batch size, and can be flexibly and independently set as a hyper-parameter.

The samples in the dictionary are progressively replaced. The current mini-batch is enqueued to the dictionary, and the oldest mini-batch in the queue is removed. The dictionary always represents a sampled subset of all data, while the extra computation of maintaining this dictionary is manageable. Moreover, removing the oldest mini-batch can be beneficial, because its encoded keys are the most outdated and thus the least consistent with the newest ones.

\paragraph{Momentum update.} Using a queue can make the dictionary large, but it also makes it intractable to update the key encoder by back-propagation (the gradient should propagate to all samples in the queue). A na\"ive solution is to copy the key encoder $f_\textrm{k}$ from the query encoder $f_\textrm{q}$, ignoring this gradient. But this solution yields poor results in experiments (Sec.~\ref{sec:imagenet}). We hypothesize that such failure is caused by the rapidly changing encoder that reduces the key representations' consistency. We propose a momentum update to address this issue.

Formally, denoting the parameters of $f_\textrm{k}$ as $\theta_\textrm{k}$ and those of $f_\textrm{q}$ as $\theta_\textrm{q}$, we update $\theta_\textrm{k}$ by:
\begin{equation}
\small
\theta_\textrm{k} \leftarrow m \theta_\textrm{k} + (1 - m) \theta_\textrm{q}.
\label{eq:moco}
\end{equation}
Here $m\in [0, 1)$ is a momentum coefficient. Only the parameters $\theta_\textrm{q}$ are updated by back-propagation. The momentum update in Eqn.(\ref{eq:moco}) makes $\theta_\textrm{k}$ evolve more smoothly than $\theta_\textrm{q}$. As a result, though the keys in the queue are encoded by different encoders (in different mini-batches), the difference among these encoders can be made small.
In experiments, a relatively large momentum (\eg, \mbox{$m$ $=$ 0.999}, our default) works much better than a smaller value (\eg, \mbox{$m$ $=$ $0.9$}), suggesting that a slowly evolving key encoder is a core to making use of a queue. 

\paragraph{Relations to previous mechanisms.} MoCo is a general mechanism for using contrastive losses. We compare it with two existing general mechanisms in Figure~\ref{fig:mechanisms}.
They exhibit different properties on the dictionary size and consistency.

The \textbf{\emph{end-to-end}} update by back-propagation is a natural mechanism 
(\eg, \cite{Hadsell2006,Oord2018,Hjelm2019,Ye2019,Bachman2019,Henaff2019}, Figure~\ref{fig:mechanisms}a). It uses samples in the current mini-batch as the dictionary, so the keys are consistently encoded (by the same set of encoder parameters).
But the dictionary size is coupled with the mini-batch size, limited by the GPU memory size. It is also challenged by large mini-batch optimization \cite{Goyal2017}.
Some recent methods \cite{Oord2018,Hjelm2019,Bachman2019} are based on pretext tasks driven by local positions, where the dictionary size can be made larger by multiple positions. But these pretext tasks may require special network designs such as patchifying the input \cite{Oord2018} or customizing the receptive field size \cite{Bachman2019}, which may complicate the transfer of these networks to downstream tasks.

Another mechanism is the \textbf{\emph{memory bank}} approach proposed by \cite{Wu2018a} (Figure~\ref{fig:mechanisms}b). A memory bank consists of the representations of all samples in the dataset. The dictionary for each mini-batch is randomly sampled from the memory bank with no back-propagation, so it can support a large dictionary size. However, the representation of a sample in the memory bank was updated when it was last seen, so the sampled keys are essentially about the encoders at \mbox{multiple} \mbox{different} steps all over the past epoch and thus are less consistent. 
A momentum update is adopted on the memory bank in \cite{Wu2018a}. Its momentum update is on the representations of the same sample, \emph{not} the \mbox{encoder}. This momentum update is \mbox{irrelevant} to our method, because MoCo does not keep track of every sample. Moreover, our method is more memory-efficient and can be trained on billion-scale data, which can be intractable for a memory bank.

Sec.~\ref{sec:exp} empirically compares these three mechanisms.

%##################################################################################################
\begin{algorithm}[t]
\caption{Pseudocode of MoCo in a PyTorch-like style.}
\label{alg:code}
\algcomment{\fontsize{7.2pt}{0em}\selectfont \texttt{bmm}: batch matrix multiplication; \texttt{mm}: matrix multiplication; \texttt{cat}: concatenation.
%\vspace{-1.em}
}
\definecolor{codeblue}{rgb}{0.25,0.5,0.5}
\lstset{
  backgroundcolor=\color{white},
  basicstyle=\fontsize{7.2pt}{7.2pt}\ttfamily\selectfont,
  columns=fullflexible,
  breaklines=true,
  captionpos=b,
  commentstyle=\fontsize{7.2pt}{7.2pt}\color{codeblue},
  keywordstyle=\fontsize{7.2pt}{7.2pt},
%  frame=tb,
}
\begin{lstlisting}[language=python]
# f_q, f_k: encoder networks for query and key
# queue: dictionary as a queue of K keys (CxK)
# m: momentum
# t: temperature

f_k.params = f_q.params  # initialize
for x in loader:  # load a minibatch x with N samples
    x_q = aug(x)  # a randomly augmented version
    x_k = aug(x)  # another randomly augmented version

    q = f_q.forward(x_q)  # queries: NxC
    k = f_k.forward(x_k)  # keys: NxC
    k = k.detach()  # no gradient to keys

    # positive logits: Nx1
    l_pos = bmm(q.view(N,1,C), k.view(N,C,1))

    # negative logits: NxK
    l_neg = mm(q.view(N,C), queue.view(C,K))

    # logits: Nx(1+K)
    logits = cat([l_pos, l_neg], dim=1)

    # contrastive loss, Eqn.(1)
    labels = zeros(N)  # positives are the 0-th
    loss = CrossEntropyLoss(logits/t, labels)

    # SGD update: query network
    loss.backward()
    update(f_q.params)

    # momentum update: key network
    f_k.params = m*f_k.params+(1-m)*f_q.params

    # update dictionary
    enqueue(queue, k)  # enqueue the current minibatch
    dequeue(queue)  # dequeue the earliest minibatch
\end{lstlisting}
\end{algorithm}
%##################################################################################################

\subsection{Pretext Task}
\label{sec:pretext}
Contrastive learning can drive a variety of pretext tasks. As the focus of this paper is not on designing a new pretext task, we use a simple one mainly following the \emph{instance discrimination} task in \cite{Wu2018a}, to which some recent works \cite{Ye2019,Bachman2019} are related. 

Following \cite{Wu2018a}, we consider a query and a key as a positive pair if they originate from the same image, and otherwise as a negative sample pair. Following \cite{Ye2019,Bachman2019}, we take two random ``views" of the same image under random data augmentation to form a positive pair.
The queries and keys are respectively encoded by their encoders, $f_\textrm{q}$ and $f_\textrm{k}$.
The encoder can be any convolutional neural network \cite{LeCun1989}.

Algorithm~\ref{alg:code} provides the pseudo-code of MoCo for this pretext task. For the current mini-batch, we encode the queries and their corresponding keys, which form the positive sample pairs. The negative samples are from the queue.

\paragraph{Technical details.}
We adopt a ResNet \cite{He2016} as the encoder, whose last fully-connected layer (after global average pooling) has a fixed-dimensional output (128-D \cite{Wu2018a}). This output vector is normalized by its L2-norm \cite{Wu2018a}. This is the representation of the query or key. The temperature $\tau$ in Eqn.(\ref{eq:infonce}) is set as 0.07 \cite{Wu2018a}.
The data augmentation setting follows \cite{Wu2018a}: a 224$\times$224-pixel crop is taken from a randomly resized image, and then undergoes random color jittering, random horizontal flip, and random grayscale conversion, all available in PyTorch's torchvision package.

\paragraph{Shuffling BN.} Our encoders $f_\textrm{q}$ and $f_\textrm{k}$ both have Batch Normalization (BN) \cite{Ioffe2015} as in the standard ResNet \cite{He2016}. In experiments, we found that using BN prevents the model from learning good representations, as similarly reported in \cite{Henaff2019} (which avoids using BN). 
The model appears to ``cheat" the pretext task and easily finds a low-loss solution. This is possibly because the intra-batch communication among samples (caused by BN) leaks information.

We resolve this problem by shuffling BN. We train with multiple GPUs and perform BN on the samples independently for each GPU (as done in common practice). 
For the key encoder $f_\textrm{k}$, we shuffle the sample order in the current mini-batch before distributing it among GPUs (and shuffle back after encoding); the sample order of the mini-batch for the query encoder $f_\textrm{q}$ is not altered. This ensures the batch statistics used to compute a query and its positive key come from two different subsets. This effectively tackles the cheating issue and allows training to benefit from BN.

We use shuffled BN in both our method and its end-to-end ablation counterpart (Figure~\ref{fig:mechanisms}a). It is irrelevant to the memory bank counterpart (Figure~\ref{fig:mechanisms}b), which does not suffer from this issue because the positive keys are from different mini-batches in the past.

\section{Experiments}
\label{sec:exp}

We study unsupervised training performed in:

\textbf{\emph{ImageNet-1M} (IN-1M)}: This is the ImageNet \cite{Deng2009} training set that has $\app$1.28 million images in 1000 classes (often called ImageNet-1K; we count the image number instead, as classes are not exploited by unsupervised learning).
This dataset is well-balanced in its class distribution, and its images generally contain iconic view of objects.

\textbf{\emph{Instagram-1B} (IG-1B)}: Following \cite{Mahajan2018}, this is a dataset of $\app$\emph{1 billion} (940M) public images from \mbox{\emph{Instagram}}. The images are from $\app$1500 hashtags \cite{Mahajan2018} that are related to the ImageNet categories. This dataset is relatively \emph{uncurated} comparing to IN-1M, and has a \emph{long-tailed}, \emph{\mbox{unbalanced}} distribution of real-world data. This dataset contains both iconic objects and scene-level images.

\paragraph{Training.} We use SGD as our optimizer. 
The SGD weight decay is 0.0001 and the SGD momentum is 0.9.
For \mbox{IN-1M}, we use a mini-batch size of 256 ($N$ in Algorithm~\ref{alg:code}) in 8 GPUs, and an initial learning rate of 0.03. We train for 200 epochs with the learning rate multiplied by 0.1 at 120 and 160 epochs \cite{Wu2018a}, taking $\app$53 hours training ResNet-50.
For IG-1B, we use a mini-batch size of 1024 in 64 GPUs, and a learning rate of 0.12 which is exponentially decayed by 0.9$\times$ after every 62.5k iterations (64M images). We train for 1.25M iterations ($\app$1.4 epochs of IG-1B), taking $\app$6 days for ResNet-50.

\subsection{Linear Classification Protocol}
\label{sec:imagenet}

We first verify our method by \emph{linear} classification on \emph{frozen} features, following a common protocol.
In this subsection we perform unsupervised pre-training on \mbox{IN-1M}. Then we freeze the features and train a supervised linear classifier (a fully-connected layer followed by softmax). We train this classifier on the global average pooling features of a ResNet, for 100 epochs. We report 1-crop, top-1 classification accuracy on the ImageNet validation set.

For this classifier, we perform a grid search and find the optimal initial learning rate is 30 and weight decay is 0 (similarly reported in \cite{Tian2019}).
These hyper-parameters perform consistently well for all ablation entries presented in this subsection. These hyper-parameter values imply that the feature distributions (\eg, magnitudes) can be substantially different from those of ImageNet supervised training, an issue we will revisit in Sec.~\ref{sec:transfer}.

\paragraph{Ablation: contrastive loss mechanisms.} We compare the three mechanisms that are illustrated in Figure~\ref{fig:mechanisms}. To focus on the effect of contrastive loss mechanisms, we implement all of them in the same pretext task as described in Sec.~\ref{sec:pretext}. We also use the same form of \mbox{InfoNCE} as the contrastive loss function, Eqn.(\ref{eq:infonce}). As such, the comparison is solely on the three mechanisms.

\begin{figure}[t]\centering
\vspace{-.5em}
\includegraphics[width=1.\linewidth]{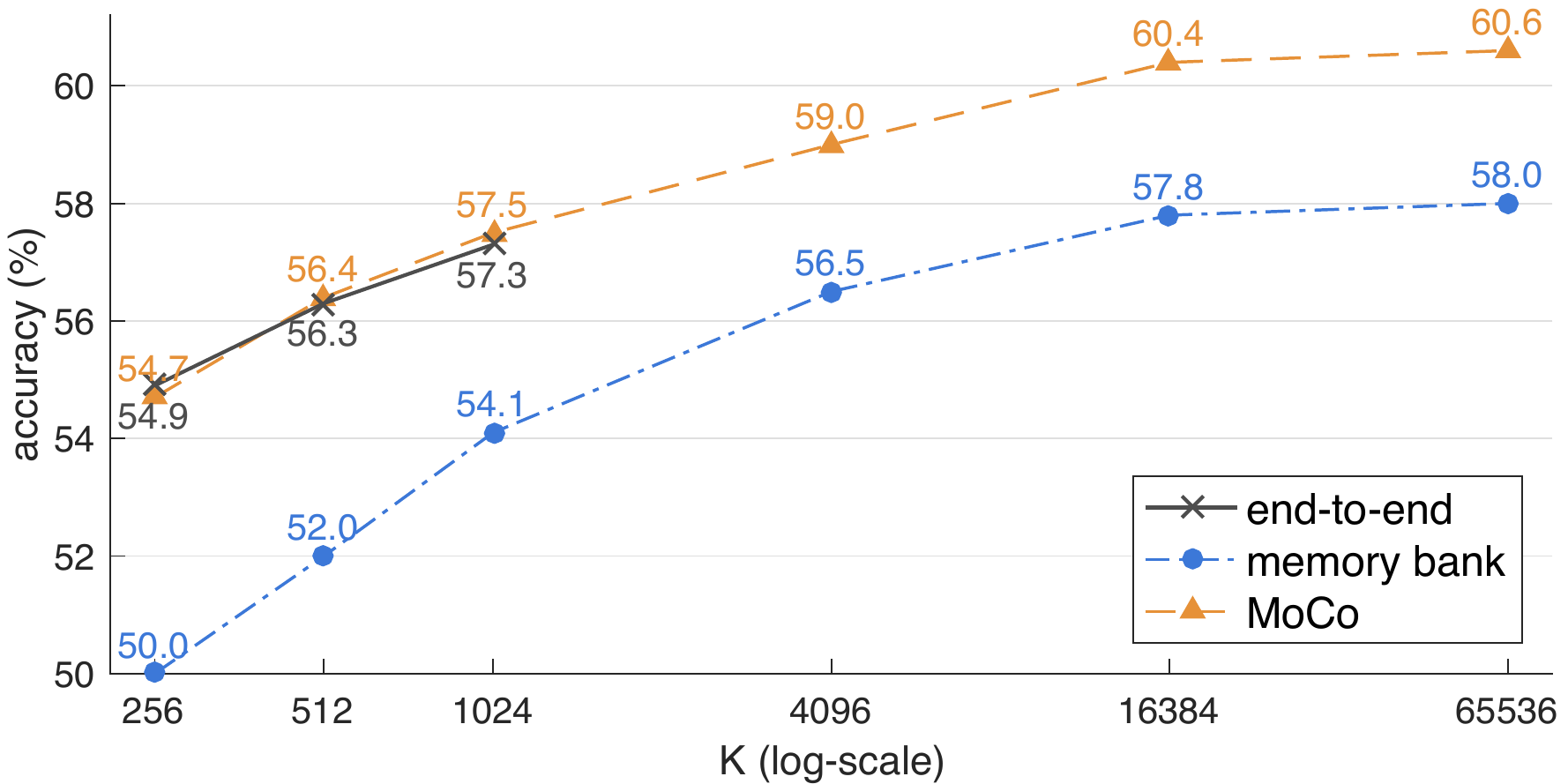}
\caption{\textbf{Comparison of three contrastive loss mechanisms} under the ImageNet linear classification protocol. We adopt the same pretext task (Sec.~\ref{sec:pretext}) and only vary the contrastive loss mechanism (Figure~\ref{fig:mechanisms}). The number of negatives is $K$ in memory bank and MoCo, and is $K{-}1$ in end-to-end (offset by one because the positive key is in the same mini-batch). The network is ResNet-50.
}
\label{fig:results_mechanisms}
\vspace{-.8em}
\end{figure}

The results are in Figure~\ref{fig:results_mechanisms}.
Overall, all three mechanisms benefit from a larger $K$.
A similar trend has been observed in \cite{Wu2018a,Tian2019} under the memory bank mechanism, while here we show that this trend is more general and can be seen in all mechanisms. These results support our motivation of building a large dictionary.

The \emph{\textbf{end-to-end}} mechanism performs similarly to MoCo when $K$ is small. However, the dictionary size is limited by the mini-batch size due to the end-to-end requirement. Here the largest mini-batch a high-end machine (8 Volta 32GB GPUs) can afford is 1024. More essentially, large mini-batch training is an open problem \cite{Goyal2017}: we found it necessary to use the linear learning rate scaling rule \cite{Goyal2017} here, without which the accuracy drops (by $\app$2\% with a 1024 mini-batch). But optimizing with a larger mini-batch is harder \cite{Goyal2017}, and it is questionable whether the trend can be extrapolated into a larger $K$ even if memory is sufficient.

The \emph{\textbf{memory bank}} \cite{Wu2018a} mechanism can support a larger dictionary size. But it is 2.6\% worse than MoCo.
This is inline with our hypothesis: the keys in the memory bank are from very different encoders all over the past epoch and they are not consistent. Note the memory bank result of 58.0\% reflects our improved implementation of \cite{Wu2018a}.\footnote{Here 58.0\% is with InfoNCE and $K$$=$65536. We reproduce 54.3\% when using NCE and $K$$=$4096 (the same as \cite{Wu2018a}), close to 54.0\% in \cite{Wu2018a}.}

%##################################################################################################
%##################################################################################################

\paragraph{Ablation: momentum.} The table below shows \mbox{ResNet-50} accuracy with different MoCo momentum values ($m$ in Eqn.(\ref{eq:moco})) used in pre-training ($K$ $=$ 4096 here) :
%##################################################################################################
\begin{center}
\vspace{-.5em}
\tablestyle{2pt}{1.2}	
\begin{tabular}{c|x{28}x{28}x{28}x{28}x{28}}
momentum $m$ & 0 & 0.9 & 0.99 & 0.999 & 0.9999 \\
\shline
accuracy (\%) & \emph{fail} & 55.2 & 57.8 & 59.0 & 58.9 \\
\end{tabular}
\vspace{-.5em}
\end{center}
%##################################################################################################
It performs reasonably well when $m$ is in 0.99 $\app$ 0.9999, showing that a slowly progressing (\ie, relatively large momentum) key encoder is beneficial.
When $m$ is too small (\eg, 0.9), the accuracy drops considerably; at the extreme of \emph{no momentum} ($m$ is 0), the training loss oscillates and fails to converge.
These results support our motivation of building a consistent dictionary.

%##################################################################################################
\begin{table}[t]
\begin{flushleft}
\vspace{-.8em}
\includegraphics[width=.95\linewidth]{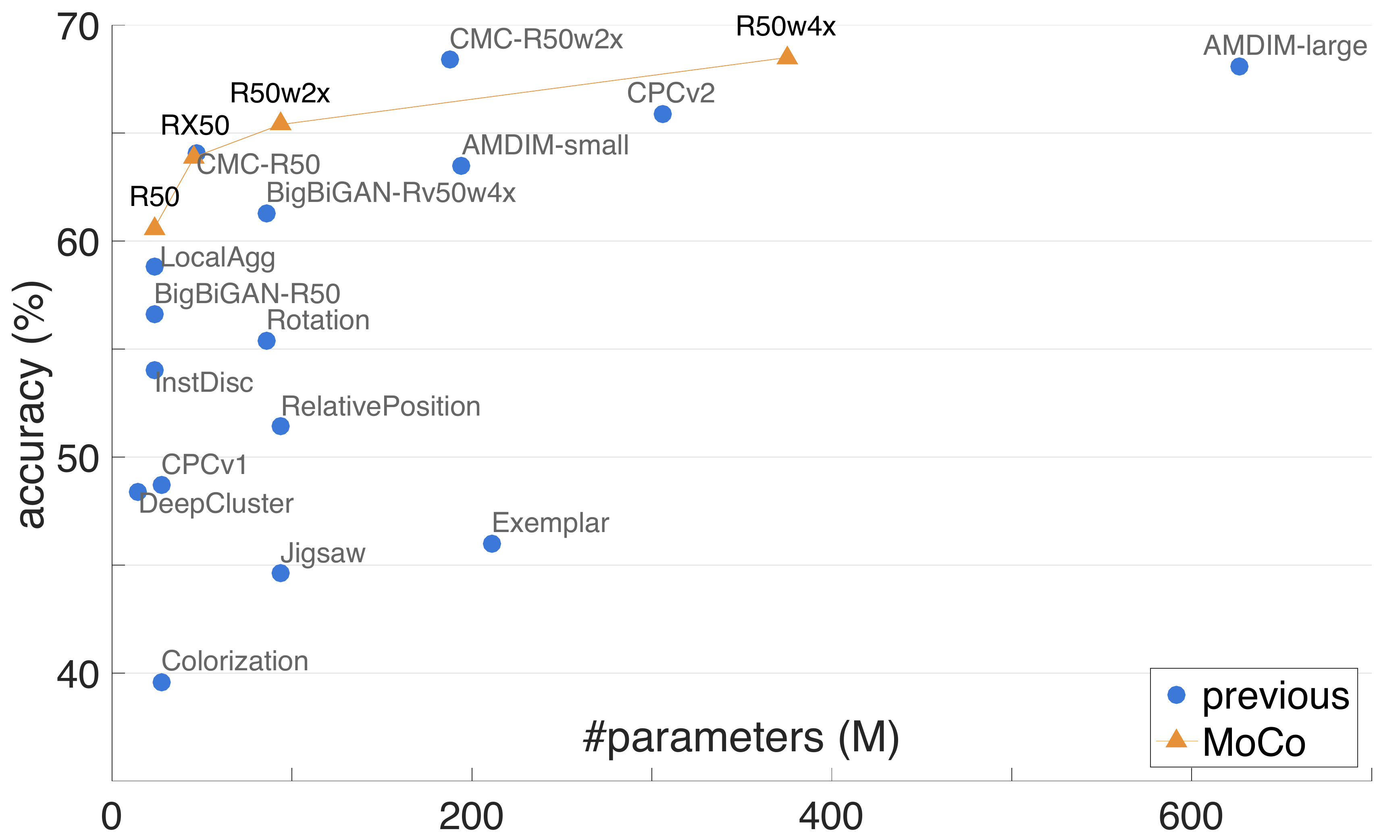}	
\end{flushleft}
\vspace{-1.em}
\centering
\tablestyle{4pt}{1.0}
\begin{tabular}{l|l|r|l}
\fontsize{7pt}{1em}\selectfont method 
& \fontsize{7pt}{1em}\selectfont architecture
& \multicolumn{1}{c|}{\fontsize{7pt}{1em}\selectfont \#params (M)}
& \fontsize{7pt}{1em}\selectfont accuracy (\%) \\
\shline
Exemplar \cite{Dosovitskiy2014} & R50w${3}{\x}$ & 211\quad~ & \quad46.0~~~\cite{Kolesnikov2019} \\
RelativePosition \cite{Doersch2015} & R50w${2}{\x}$ & 94\quad~ & \quad51.4~~~\cite{Kolesnikov2019} \\
Jigsaw \cite{Noroozi2016} & R50w${2}{\x}$ & 94\quad~ & \quad44.6~~~\cite{Kolesnikov2019} \\
Rotation \cite{Gidaris2018} & Rv50w${4}{\x}$ & 86\quad~ & \quad55.4~~~\cite{Kolesnikov2019} \\
Colorization \cite{Zhang2016} & R101$^*$ & 28\quad~ & \quad39.6~~~\cite{Doersch2017} \\
DeepCluster \cite{Caron2018} & VGG \cite{Simonyan2015} & 15\quad~ & \quad48.4~~~\cite{Caron2019} \\
BigBiGAN \cite{Donahue2019} & R50 & 24\quad~ & \quad56.6  \\
 & Rv50w${4}{\x}$ & 86\quad~ & \quad61.3  \\
\hline
\multicolumn{4}{l}{\emph{methods based on contrastive learning follow:}} \\
\hline
InstDisc \cite{Wu2018a} & R50 & 24\quad~ & \quad54.0 \\
LocalAgg \cite{Zhuang2019} & R50 & 24\quad~ & \quad58.8 \\
CPC v1 \cite{Oord2018} & R101$^*$ & 28\quad~ & \quad48.7 \\
CPC v2 \cite{Henaff2019} & R170$^*_\text{wider}$ & 303\quad~ & \quad65.9 \\
CMC \cite{Tian2019} & R50$_\text{L+ab}$  & 47\quad~ & \quad64.1$^\dagger$ \\
	& R50w${2}{\x}$$_\text{L+ab}$ & 188\quad~ & \quad68.4$^\dagger$ \\
AMDIM \cite{Bachman2019} & AMDIM$_\text{small}$ & 194\quad~ & \quad63.5$^\dagger$ \\
 & AMDIM$_\text{large}$ & 626\quad~ & \quad68.1$^\dagger$ \\
\hline
\textbf{MoCo} & R50 & 24\quad~ &  \quad60.6 \\
 & RX50 & 46\quad~ &  \quad63.9 \\
 & R50w${2}{\x}$ & 94\quad~ &  \quad65.4 \\
 & R50w${4}{\x}$ & 375\quad~ &  \quad\textbf{68.6} \\
\end{tabular}
\vspace{.1em}
% -------------------------
\caption{\textbf{Comparison under the linear classification protocol on ImageNet}. The figure visualizes the table.
All are reported as \mbox{unsupervised} pre-training on the ImageNet-1M training set, followed by supervised linear classification trained on frozen features, evaluated on the validation set.
The parameter counts are those of the feature extractors.
We compare with improved re-implementations if available (referenced after the numbers).
\newline
{\footnotesize
Notations: R101$^*$/R170$^*$ is ResNet-101/170 with the last residual stage removed \cite{Doersch2017,Oord2018,Henaff2019}, and R170 is made wider \cite{Henaff2019}; Rv50 is a reversible net \cite{Gomez2017}, RX50 is ResNeXt-50-32${\x}$8d \cite{Xie2017}.
}
\newline
{\fontsize{7.2pt}{0pt}\selectfont{
$^\dagger$: \emph{Pre-training uses FastAutoAugment \cite{Lim2019} that is supervised by ImageNet labels.
}}}
} % end caption
\label{tab:results_imgnet}
\vspace{-1.5em}
\end{table}
%##################################################################################################

\paragraph{Comparison with previous results.}

Previous unsupervised learning methods can differ substantially in model sizes. 
For a fair and comprehensive comparison, we report \textbf{\emph{accuracy} \vs \emph{\#parameters}}\footnote{Parameters are of the \emph{feature extractor}: \eg, we do not count the parameters of \emph{conv}$_\text{x}$ if \emph{conv}$_\text{x}$ is not included in linear classification.}
trade-offs.
Besides \mbox{ResNet-50} (R50) \cite{He2016}, we also report its variants that are ${2}{\x}$ and ${4}{\x}$ wider (more channels), following \cite{Kolesnikov2019}.\footnote{Our w${2}{\x}$ and w${4}{\x}$ models correspond to the ``$\times$8" and ``$\times$16'' cases in \cite{Kolesnikov2019}, because the standard-sized ResNet is referred to as ``$\times$4" in \cite{Kolesnikov2019}.}
We set \mbox{$K$ $=$ 65536} and \mbox{$m$ $=$ 0.999}. Table~\ref{tab:results_imgnet} is the comparison.

MoCo with R50 performs competitively and achieves 60.6\% accuracy, better than all competitors of similar model sizes ($\app$24M).
MoCo benefits from larger models and achieves 68.6\% accuracy with R50w${4}{\x}$.

Notably, we achieve competitive results using a \emph{standard} ResNet-50 and require no specific architecture designs, \eg, patchified inputs \cite{Oord2018,Henaff2019}, carefully tailored receptive fields \cite{Bachman2019}, or combining two networks \cite{Tian2019}. By using an architecture that is not customized for the pretext task, it is easier to transfer features to a variety of visual tasks and make comparisons, studied in the next subsection.

This paper's focus is on a mechanism for general contrastive learning; we do not explore orthogonal factors (such as specific pretext tasks) that may further improve accuracy. As an example, ``MoCo v2" \cite{Chen2020a}, an extension of a preliminary version of this manuscript, achieves 71.1\% accuracy with R50 (up from 60.6\%), given small changes on the data augmentation and output projection head \cite{Chen2020}. We believe that this additional result shows the generality and robustness of the MoCo framework.

%##################################################################################################

\definecolor{Gray}{gray}{0.5}

\newcommand{\randinit}{\tablestyle{1pt}{1} \begin{tabular}{z{21}y{26}} \multicolumn{2}{c}{\demph{random init.}} \end{tabular}}
\newcommand{\mocoimgnet}{\tablestyle{0pt}{1} \begin{tabular}{z{21}y{26}} \textbf{MoCo} & ~~IN-1M \end{tabular}}
\newcommand{\mocoins}{\tablestyle{0pt}{1} \begin{tabular}{z{21}y{26}} \textbf{MoCo} & ~~IG-1B \end{tabular}}
\newcommand{\supimgnet}{\tablestyle{0pt}{1} \begin{tabular}{z{21}y{26}} super. & ~~IN-1M \end{tabular}}
% ------------------------------------------------

\newcommand{\demph}[1]{\textcolor{Gray}{#1}}
\newcommand{\std}[1]{{\fontsize{5pt}{1em}\selectfont ~~$_\pm$$_{\text{#1}}$}}

\definecolor{Highlight}{HTML}{39b54a}  % green

\renewcommand{\hl}[1]{\textcolor{Highlight}{#1}}

\newcommand{\res}[3]{
\tablestyle{1pt}{1}
\begin{tabular}{z{16}y{18}}
{#1} &
\fontsize{7.5pt}{1em}\selectfont{~(${#2}${#3})}
\end{tabular}}

\newcommand{\reshl}[3]{
\tablestyle{1pt}{1} 
\begin{tabular}{z{16}y{18}}
{#1} &
\fontsize{7.5pt}{1em}\selectfont{~\hl{(${#2}$\textbf{#3})}}
\end{tabular}}

\newcommand{\resrand}[2]{\tablestyle{1pt}{1} \begin{tabular}{z{16}y{18}} \demph{#1} & {} \end{tabular}}
\newcommand{\ressup}[2]{\tablestyle{1pt}{1} \begin{tabular}{z{16}y{18}} {#1} & {} \end{tabular}}

\begin{table}[t]
\small
\vspace{-1.em}
\centering
% ------------------------------------------------
\subfloat[Faster R-CNN, R50-\textbf{dilated-C5}]{
\tablestyle{1pt}{1.0}
\begin{tabular}{x{56}|x{54}|x{54}x{54}c}
pre-train &
AP$_\text{50}$ &
AP &
AP$_\text{75}$ & \\ 
\shline
\randinit & \resrand{64.4}{} & \resrand{37.9}{} & \resrand{38.6}{} & \\
\supimgnet  & \ressup{81.4}{} & \ressup{54.0}{} & \ressup{59.1}{} & \\
\hline
\mocoimgnet & \res{81.1}{-}{0.3} & \reshl{54.6}{+}{0.6} & \reshl{59.9}{+}{0.8} & \\
\mocoins & \res{81.6}{+}{0.2} & \reshl{55.5}{+}{1.5} & \reshl{61.2}{+}{2.1} & \\
\end{tabular}	
} % end of subfloat
% ------------------------------------------------
\\
\vspace{-.5em}
% ------------------------------------------------
\subfloat[Faster R-CNN, R50-\textbf{C4}]{
\tablestyle{1pt}{1.0}
\begin{tabular}{x{56}|x{54}|x{54}x{54}c}
pre-train &
AP$_\text{50}$ &
AP &
AP$_\text{75}$ & \\ 
\shline
\randinit & \resrand{60.2}{} & \resrand{33.8}{} & \resrand{33.1}{} & \\
\supimgnet & \ressup{81.3}{} & \ressup{53.5}{} & \ressup{58.8}{} & \\
\hline
\mocoimgnet & \res{81.5}{+}{0.2} & \reshl{55.9}{+}{2.4} & \reshl{62.6}{+}{3.8} & \\
\mocoins & \reshl{82.2}{+}{0.9} & \reshl{57.2}{+}{3.7} & \reshl{63.7}{+}{4.9} & \\
\end{tabular}	
} % end of subfloat
% ------------------------------------------------
\vspace{.3em}
\caption{\textbf{Object detection fine-tuned on PASCAL VOC} {\texttt{trainval07+12}}. Evaluation is on \textnormal{\texttt{test2007}}: AP$_\text{50}$ (default VOC metric), AP (COCO-style), and AP$_\text{75}$, averaged over 5 trials.
All are fine-tuned for 24k iterations ($\app$23 epochs). In the brackets are the gaps to the ImageNet supervised pre-training counterpart. In green are the gaps of at least {\fontsize{8pt}{1em}\selectfont \hl{${+}$\textbf{0.5}}} point.
}
\label{tab:voc}
\vspace{-.3em}
\end{table}
%##################################################################################################

%%##################################################################################################
\begin{table}[t]
\vspace{-.5em}
\tablestyle{1.8pt}{1.05}
\begin{tabular}{x{48}|x{26}|x{26}x{26}|x{26}|x{26}x{26}c}
 &
\multicolumn{3}{c|}{R50-dilated-C5} & 
\multicolumn{3}{c}{R50-C4} & \\
pre-train &
AP$_\text{50}$ &
AP &
AP$_\text{75}$ &
AP$_\text{50}$ &
AP &
AP$_\text{75}$ & \\ 
\shline
end-to-end &
79.2 & 52.0 & 56.6 & % e2e dC5
80.4 & 54.6 & 60.3 & % e2e C4
\\
memory bank &
79.8 & 52.9 & 57.9 & % mbank dC5
80.6 & 54.9 & 60.6 % mbank C4
\\
\hline
\textbf{MoCo} &
\textbf{81.1} & \textbf{54.6} & \textbf{59.9} & % moco dC5 
\textbf{81.5} & \textbf{55.9} & \textbf{62.6} \\ % moco C4
\end{tabular}
\vspace{.3em}
\caption{\textbf{Comparison of three contrastive loss mechanisms} on PASCAL VOC object detection, fine-tuned on \texttt{trainval07+12} and evaluated on \texttt{test2007} (averages over 5 trials). All models are implemented by us (Figure~\ref{fig:results_mechanisms}), pre-trained on IN-1M, and fine-tuned using the same settings as in Table~\ref{tab:voc}.}
\label{tab:voc_mechanisms}
\vspace{-1.em}
\end{table}
%%##################################################################################################

%##################################################################################################
\renewcommand{\res}[3]{
\tablestyle{1pt}{1}
\begin{tabular}{z{14}y{18}}
{#1} &
\fontsize{7.5pt}{1em}\selectfont{(${#2}${#3})}
\end{tabular}}

\renewcommand{\reshl}[3]{
\tablestyle{1pt}{1}
\begin{tabular}{z{14}y{18}}
{#1} &
\fontsize{7.5pt}{1em}\selectfont{\textcolor{Highlight}{(${#2}$\textbf{#3})}} 
\end{tabular}}

\renewcommand{\resrand}[2]{\tablestyle{1pt}{1} \begin{tabular}{z{14}y{18}} \demph{#1} & {} \end{tabular}}
\renewcommand{\ressup}[2]{\tablestyle{1pt}{1} \begin{tabular}{z{14}y{18}} {#1} & {} \end{tabular}}

\newcommand{\tagfont}[1]{{\fontsize{7pt}{1em}\selectfont #1}}

\begin{table*}[t]
\vspace{-1.em}
\small
\tablestyle{1.8pt}{1.05}
\begin{tabular}{ry{32}|y{52}y{52}y{52}y{52}y{52}|y{52}|y{52}y{48}c}
& &
\multicolumn{5}{c|}{AP$_\text{50}$} &
\multicolumn{1}{c|}{AP} &
\multicolumn{2}{c}{AP$_\text{75}$}
\\
\multicolumn{2}{c|}{pre-train} &
\multicolumn{1}{c}{RelPos, by \cite{Doersch2017}} &
\multicolumn{1}{c}{Multi-task \cite{Doersch2017}} &
\multicolumn{1}{c}{Jigsaw, by \cite{Goyal2019}} &
\multicolumn{1}{c}{LocalAgg \cite{Zhuang2019}} &
\multicolumn{1}{c|}{\textbf{MoCo}} &
\multicolumn{1}{c|}{\textbf{MoCo}} &
\multicolumn{1}{c}{Multi-task \cite{Doersch2017}} &
\multicolumn{1}{c}{\textbf{MoCo}} &
\\
\shline
\tagfont{super.} &
\tagfont{IN-1M} &
\ressup{74.2}{} &  % relpos from multi-task AP50
\ressup{74.2}{} &  % multi-task AP50
\ressup{70.5}{} &  % jigsaw AP50
\ressup{74.6}{} &  % localagg AP50
\ressup{74.4}{} &  % ours AP50
\ressup{42.4}{} &  % ours AP
\ressup{44.3}{} &  % multi-task AP75
\ressup{42.7}{} &  % ours AP75
\\
\hline
\tagfont{unsup.} &
\tagfont{IN-1M} &
\res{66.8}{-}{7.4} &
\res{70.5}{-}{3.7} &
\res{61.4}{-}{9.1} &
\res{69.1}{-}{5.5} &
\reshl{74.9}{+}{0.5} &
\reshl{46.6}{+}{4.2} &
\res{43.9}{-}{0.4} &
\reshl{50.1}{+}{7.4} &
\\
\tagfont{unsup.} &
\tagfont{IN-14M} &
\ressup{-}{} &
\ressup{-}{} & 
\res{69.2}{-}{1.3} &
\ressup{-}{} &
\reshl{75.2}{+}{0.8} &
\reshl{46.9}{+}{4.5} &
\ressup{-}{} &
\reshl{50.2}{+}{7.5} &
\\
\tagfont{unsup.} &
{\fontsize{5.8pt}{1em}\selectfont YFCC-100M} &
\ressup{-}{} &
\ressup{-}{} & 
\res{66.6}{-}{3.9} &
\ressup{-}{} &
\res{74.7}{+}{0.3} &
\reshl{45.9}{+}{3.5} &
\ressup{-}{} &
\reshl{49.0}{+}{6.3} &
\\
\tagfont{unsup.} &
\tagfont{IG-1B} &
\ressup{-}{} &
\ressup{-}{} & 
\ressup{-}{} &
\ressup{-}{} &
\reshl{75.6}{+}{1.2} &
\reshl{47.6}{+}{5.2} &
\ressup{-}{} &
\reshl{51.7}{+}{9.0} &
\end{tabular}
%\vspace{.05em}
\caption{
\textbf{Comparison with previous methods on object detection fine-tuned on PASCAL VOC} {\texttt{trainval2007}}. Evaluation is on \textnormal{\texttt{test2007}}.
The ImageNet supervised counterparts are from the respective papers, and are reported as having \emph{the same structure} as the respective unsupervised pre-training counterparts.
All entries are based on the C4 backbone.
The models in \cite{Doersch2017} are R101 v2 \cite{He2016a}, and others are R50.
The RelPos (relative position) \cite{Doersch2015} result is the best single-task case in the Multi-task paper \cite{Doersch2017}.
The Jigsaw \cite{Noroozi2016} result is from the ResNet-based implementation in \cite{Goyal2019}.
Our results are with 9k-iteration fine-tuning, averaged over 5 trials.
In the brackets are the gaps to the ImageNet supervised pre-training counterpart. In green are the gaps of at least {\fontsize{8pt}{1em}\selectfont \hl{${+}$\textbf{0.5}}} point.
}
\label{tab:voc07}
\vspace{-1.em}
\end{table*}
%##################################################################################################

\subsection{Transferring Features}
\label{sec:transfer}

A main goal of unsupervised learning is to \emph{learn features that are transferrable}. ImageNet supervised pre-training is most influential when serving as the initialization for fine-tuning in downstream tasks (\eg, \cite{Girshick2014,Girshick2015,Long2015,Ren2015}).
Next we compare MoCo with ImageNet supervised pre-training, transferred to various tasks including PASCAL VOC \cite{Everingham2010}, COCO \cite{Lin2014}, \etc.
As prerequisites, we discuss two important issues involved \cite{He2019}: normalization and schedules.

\paragraph{Normalization.} As noted in Sec.~\ref{sec:imagenet}, features produced by unsupervised pre-training can have different distributions compared with ImageNet supervised pre-training. 
But a system for a downstream task often has hyper-parameters (\eg, learning rates) selected for supervised pre-training.
To relieve this problem, we adopt \emph{feature normalization} during fine-tuning: we fine-tune with BN that is trained (and synchronized across GPUs \cite{Peng2018}), instead of freezing it by an affine layer \cite{He2016}. We also use BN in the newly initialized layers (\eg, FPN \cite{Lin2017}), which helps calibrate magnitudes.

We perform normalization when fine-tuning supervised and unsupervised pre-training models. MoCo \ul{\emph{uses the same hyper-parameters}} as the ImageNet supervised counterpart.

\paragraph{Schedules.} If the fine-tuning schedule is long enough, training detectors from random initialization can be strong baselines, and can match the ImageNet supervised counterpart on COCO \cite{He2019}. Our goal is to investigate \emph{transferability} of features, so our experiments are on controlled schedules, \eg, the 1$\x$ ($\app$12 epochs) or 2$\x$ schedules \cite{Detectron2018} for COCO, in contrast to 6$\x$$\app$9$\x$ in \cite{He2019}. On smaller datasets like VOC, training longer may not catch up \cite{He2019}.

Nonetheless, in our fine-tuning, MoCo \ul{\emph{uses the same schedule}} as the ImageNet supervised counterpart, and random initialization results are provided as references.

\vspace{.5em}
Put together, our fine-tuning uses the same setting as the supervised pre-training counterpart. This may place MoCo at a \emph{disadvantage}. Even so, MoCo is competitive. Doing so also makes it feasible to present comparisons on multiple datasets/tasks, without extra hyper-parameter search.

\vspace{-.5em}
\subsubsection{PASCAL VOC Object Detection}

\paragraph{Setup.}
The detector is Faster R-CNN \cite{Ren2015} with a backbone of R50-dilated-C5 or R50-C4 \cite{He2017} (details in \appdx), with BN tuned, implemented in \cite{Wu2019}.
We fine-tune all layers end-to-end. The image scale is [480, 800] pixels during training and 800 at inference. The same setup is used for all entries, including the supervised pre-training baseline.
We evaluate the default VOC metric of AP$_\text{50}$ (\ie, IoU threshold is 50\%) and the more stringent metrics of COCO-style AP and AP$_\text{75}$. Evaluation is on the VOC \texttt{test2007} set.

\paragraph{Ablation: backbones.}
Table~\ref{tab:voc} shows the results fine-tuned on \texttt{trainval07+12} ($\app$16.5k images).
For R50-dilated-C5 (Table~\ref{tab:voc}a), MoCo pre-trained on IN-1M is \mbox{comparable} to the supervised pre-training counterpart, and MoCo pre-trained on IG-1B \emph{surpasses} it.
For R50-C4 (Table~\ref{tab:voc}b), MoCo with IN-1M or IG-1B is \emph{better} than the supervised counterpart: up to +\textbf{0.9} AP$_\text{50}$, +\textbf{3.7} AP, and +\textbf{4.9} AP$_\text{75}$.

Interestingly, the transferring accuracy depends on the detector structure. For the C4 backbone, by default used in existing ResNet-based results \cite{Doersch2017,Wu2018a,Goyal2019,Zhuang2019}, the advantage of unsupervised pre-training is larger. The relation between pre-training \vs detector structures has been veiled in the past, and should be a factor under consideration.

\paragraph{Ablation: contrastive loss mechanisms.}
We point out that these results are partially because we establish solid detection baselines for contrastive learning. 
To pin-point the gain that is \emph{solely} contributed by using the MoCo mechanism in contrastive learning, we fine-tune the models pre-trained with the end-to-end or memory bank mechanism, both implemented by us (\ie, the best ones in Figure~\ref{fig:results_mechanisms}), using the same fine-tuning setting as MoCo.

These competitors perform decently (Table~\ref{tab:voc_mechanisms}). Their AP and AP$_\text{75}$ with the C4 backbone are also higher than the ImageNet supervised counterpart's, \cf Table~\ref{tab:voc}b, but other metrics are lower.
They are worse than MoCo in all metrics. This shows the benefits of MoCo.  In addition, how to train these competitors in larger-scale data is an open question, and they may not benefit from IG-1B.

\paragraph{Comparison with previous results.} Following the competitors, we fine-tune on \texttt{trainval2007} ($\app$5k images) using the C4 backbone. The comparison is in Table~\ref{tab:voc07}.

For the AP$_\text{50}$ metric, \emph{no} previous method can catch up with its respective supervised pre-training counterpart. MoCo pre-trained on any of IN-1M, IN-14M (full ImageNet), YFCC-100M \cite{Thomee2016}, and IG-1B can \emph{outperform} the supervised baseline. Large gains are seen in the more stringent metrics: up to +\textbf{5.2}~AP and +\textbf{9.0}~AP$_\text{75}$. These gains are larger than the gains seen in \texttt{trainval07+12} (Table~\ref{tab:voc}b).

%##################################################################################################
\newcommand{\cres}[1]{#1}
\newcommand{\cresrand}[1]{#1}

\newcommand{\cgap}[2]{
\fontsize{6pt}{1em}\selectfont{(${#1}${#2})}
}
\newcommand{\cgaphl}[2]{
\fontsize{6pt}{1em}\selectfont{\textcolor{Highlight}{(${#1}$\textbf{#2})}}
}

\renewcommand{\randinit}{\tablestyle{1pt}{1} \begin{tabular}{z{21}y{26}} \multicolumn{2}{c}{random init.} \end{tabular}}

\begin{table*}[t]
\vspace{-1.em}
\small
% ------------------------------------------------
\hspace{-1.5em}
\resizebox{1.05\linewidth}{!}{
\subfloat[Mask R-CNN, R50-\textbf{FPN}, \textbf{1$\x$} schedule]{
\tablestyle{.8pt}{1.05}
\begin{tabular}{cr|
z{17}y{18}
z{17}y{18}
z{17}y{18}|
z{17}y{18}
z{17}y{18}
z{17}y{18}c
}
pre-train & ~ &
\multicolumn{2}{c}{\apbbox{~}} &
\multicolumn{2}{c}{\apbbox{50}} &
\multicolumn{2}{c|}{\apbbox{75}} &
\multicolumn{2}{c}{\apmask{~}} &
\multicolumn{2}{c}{\apmask{50}} &
\multicolumn{2}{c}{\apmask{75}} &\\
\shline
\demph{\randinit} & ~ &
\demph{31.0} & ~ & \demph{49.5} & ~ & \demph{33.2} & ~ &
\demph{28.5} & ~ & \demph{46.8} & ~ & \demph{30.4} & ~ & \\
\supimgnet & ~ &
38.9 & ~ & 59.6 & ~ & 42.7 & ~ &
35.4 & ~ & 56.5 & ~ & 38.1 & ~ & \\
\hline
\mocoimgnet & ~ &
38.5 & \cgap{-}{0.4} & 58.9 & \cgap{-}{0.7} & 42.0 & \cgap{-}{0.7} &
35.1 & \cgap{-}{0.3} & 55.9 & \cgap{-}{0.6} & 37.7 & \cgap{-}{0.4} & \\
\mocoins & ~ &
38.9 & \cgap{{\transparent{0}+}}{0.0} & 59.4 & \cgap{-}{0.2} & 42.3 & \cgap{-}{0.4} &
35.4 & \cgap{{\transparent{0}+}}{0.0} & 56.5 & \cgap{{\transparent{0}+}}{0.0} & 37.9 & \cgap{-}{0.2} & \\
\end{tabular}	
}  % end of subfloat
% ------------------------------------------------

% ------------------------------------------------
\subfloat[Mask R-CNN, R50-\textbf{FPN}, \textbf{2$\x$} schedule]{
\tablestyle{.8pt}{1.05}
\begin{tabular}{
z{17}y{18}
z{17}y{18}
z{17}y{18}|
z{17}y{18}
z{17}y{18}
z{17}y{18}c
}
%	pre-train &
\multicolumn{2}{c}{\apbbox{~}} &
\multicolumn{2}{c}{\apbbox{50}} &
\multicolumn{2}{c|}{\apbbox{75}} &
\multicolumn{2}{c}{\apmask{~}} &
\multicolumn{2}{c}{\apmask{50}} &
\multicolumn{2}{c}{\apmask{75}} &\\
\shline
%\demph{\randinit} &
\demph{36.7} & ~  & \demph{56.7} & ~  & \demph{40.0} & ~  &
\demph{33.7} & ~  & \demph{53.8} & ~  & \demph{35.9} & ~  & \\
%\supimgnet &
40.6 & ~ & 61.3 & ~ & 44.4 & ~ &
36.8 & ~ & 58.1 & ~ & 39.5 & ~ & \\
\hline
%\mocoimgnet &
40.8 & \cgap{+}{0.2} & 61.6 & \cgap{+}{0.3} & 44.7 & \cgap{+}{0.3} &
36.9 & \cgap{+}{0.1} & 58.4 & \cgap{+}{0.3} & 39.7 & \cgap{+}{0.2} & \\
%\mocoins &
41.1 & \cgaphl{+}{0.5} & 61.8 & \cgaphl{+}{0.5} & 45.1 & \cgaphl{+}{0.7} & 
37.4 & \cgaphl{+}{0.6} & 59.1 & \cgaphl{+}{1.0} & 40.2 & \cgaphl{+}{0.7} & \\
\end{tabular}	
}  % end of subfloat
% ------------------------------------------------
}  % end of resizebox

\vspace{-.5em}
\hspace{-1.5em}
\resizebox{1.05\linewidth}{!}{
\subfloat[Mask R-CNN, R50-\textbf{C4}, \textbf{1$\x$} schedule]{
\tablestyle{.8pt}{1.05}
\begin{tabular}{cr|
z{17}y{18}
z{17}y{18}
z{17}y{18}|
z{17}y{18}
z{17}y{18}
z{17}y{18}c
}
pre-train & ~ &
\multicolumn{2}{c}{\apbbox{~}} &
\multicolumn{2}{c}{\apbbox{50}} &
\multicolumn{2}{c|}{\apbbox{75}} &
\multicolumn{2}{c}{\apmask{~}} &
\multicolumn{2}{c}{\apmask{50}} &
\multicolumn{2}{c}{\apmask{75}} &\\
\shline
\demph{\randinit} & ~ &
\demph{26.4} & ~ & \demph{44.0} & ~ & \demph{27.8} & ~ &
\demph{29.3} & ~ & \demph{46.9} & ~ & \demph{30.8} & ~ & \\
\supimgnet & ~ &
38.2 & ~ & 58.2 & ~ & 41.2 & ~ &
33.3 & ~ & 54.7 & ~ & 35.2 & ~ & \\
\hline
\mocoimgnet & ~ &
38.5 & \cgap{+}{0.3} & 58.3 & \cgap{+}{0.1} & 41.6 & \cgap{+}{0.4} &
33.6 & \cgap{+}{0.3} & 54.8 & \cgap{+}{0.1} & 35.6 & \cgap{+}{0.4} & \\
\mocoins & ~ &
39.1 & \cgaphl{+}{0.9} & 58.7 & \cgaphl{+}{0.5} & 42.2 & \cgaphl{+}{1.0} &
34.1 & \cgaphl{+}{0.8} & 55.4 & \cgaphl{+}{0.7} & 36.4 & \cgaphl{+}{1.2} & \\
\end{tabular}	
}  % end of subfloat
% ------------------------------------------------

% ------------------------------------------------
\subfloat[Mask R-CNN, R50-\textbf{C4}, \textbf{2$\x$} schedule]{
\tablestyle{.8pt}{1.05}
\begin{tabular}{
z{17}y{18}
z{17}y{18}
z{17}y{18}|
z{17}y{18}
z{17}y{18}
z{17}y{18}c
}
%	pre-train &
\multicolumn{2}{c}{\apbbox{~}} &
\multicolumn{2}{c}{\apbbox{50}} &
\multicolumn{2}{c|}{\apbbox{75}} &
\multicolumn{2}{c}{\apmask{~}} &
\multicolumn{2}{c}{\apmask{50}} &
\multicolumn{2}{c}{\apmask{75}} &\\
\shline
%\demph{\randinit} &
\demph{35.6} & ~  & \demph{54.6} & ~  & \demph{38.2} & ~  &
\demph{31.4} & ~  & \demph{51.5} & ~  & \demph{33.5} & ~  & \\
%\supimgnet &
40.0 & ~ & 59.9 & ~ & 43.1 & ~ &
34.7 & ~ & 56.5 & ~ & 36.9 & ~ & \\
\hline
%\mocoimgnet &
40.7 & \cgaphl{+}{0.7} & 60.5 & \cgaphl{+}{0.6} & 44.1 & \cgaphl{+}{1.0} &
35.4 & \cgaphl{+}{0.7} & 57.3 & \cgaphl{+}{0.8} & 37.6 & \cgaphl{+}{0.7} & \\
%\mocoins &
41.1 & \cgaphl{+}{1.1} & 60.7 & \cgaphl{+}{0.8} & 44.8 & \cgaphl{+}{1.7} &
35.6 & \cgaphl{+}{0.9} & 57.4 & \cgaphl{+}{0.9} & 38.1 & \cgaphl{+}{1.2} & \\
\end{tabular}	
}  % end of subfloat
% ------------------------------------------------
}  % end of resizebox
% ------------------------------------------------
%\vspace{.1em}
\caption{\textbf{Object detection and instance segmentation fine-tuned on COCO}: bounding-box AP (\apbbox{}) and mask AP (\apmask{}) evaluated on \texttt{val2017}. In the brackets are the gaps to the ImageNet supervised pre-training counterpart. In green are the gaps of at least {\fontsize{8pt}{1em}\selectfont \hl{${+}$\textbf{0.5}}} point.
}
\label{tab:coco}
\vspace{-1.em}
\end{table*}
%##################################################################################################

\vspace{-.5em}
\subsubsection{COCO Object Detection and Segmentation}

\paragraph{Setup.}
The model is Mask R-CNN \cite{He2017} with the FPN \cite{Lin2017} or C4 backbone, with BN tuned, implemented in \cite{Wu2019}.
The image scale is in [640, 800] pixels during training and is 800 at inference. We fine-tune all layers end-to-end.
We fine-tune on the \texttt{train2017} set ($\app$118k images) and evaluate on \texttt{val2017}.
The schedule is the default 1$\x$ or 2$\x$ in \cite{Detectron2018}.

\paragraph{Results.}

Table~\ref{tab:coco} shows the results on COCO with the FPN (Table~\ref{tab:coco}a, b) and C4 (Table~\ref{tab:coco}c, d) backbones.
With the 1$\x$ schedule, all models (including the ImageNet supervised counterparts) are heavily under-trained, as indicated by the $\app$2 points gaps to the 2$\x$ schedule cases. 
With the 2$\x$ schedule, MoCo is \emph{better} than its ImageNet supervised counterpart in all metrics in both backbones.

%##################################################################################################

\begin{table}[t]
\centering
%% coco kp
\begin{raggedleft}
\tablestyle{2pt}{1.1}
\begin{tabular}{ry{28}|x{48}x{48}x{48} c}
\multicolumn{2}{c|}{} &
\multicolumn{3}{c}{\fontsize{7.5pt}{1em}\selectfont \textbf{COCO keypoint detection}} & 
\\
\multicolumn{2}{c|}{pre-train} &
\apkp{~} & \apkp{50} & \apkp{75} &  % coco kp
\\
\shline
\multicolumn{2}{c|}{\demph{random init.}}
& \resrand{65.9}{} & \resrand{86.5}{} & \resrand{71.7}{}  % coco kp
& \\
super. & IN-1M
& \ressup{65.8}{} & \ressup{86.9}{} & \ressup{71.9}{}  % coco kp
& \\
\hline
\textbf{MoCo} & IN-1M
& \reshl{66.8}{+}{1.0} & \reshl{87.4}{+}{0.5} & \reshl{72.5}{+}{0.6}  % coco kp
& \\
\textbf{MoCo} & IG-1B
& \reshl{66.9}{+}{1.1} & \reshl{87.8}{+}{0.9} & \reshl{73.0}{+}{1.1}  % coco kp
& \\
\end{tabular}
\end{raggedleft}
\vspace{.3em}

%% coco dp
\begin{raggedleft}
\tablestyle{2pt}{1.1}
\begin{tabular}{ry{28}|x{48}x{48}x{48} c}
\multicolumn{2}{c|}{} &
\multicolumn{3}{c}{\fontsize{7.5pt}{1em}\selectfont \textbf{COCO dense pose estimation}} & 
\\
\multicolumn{2}{c|}{pre-train} &
\apdp{~} & \apdp{50} & \apdp{75} &  % coco pose
\\
\shline
\multicolumn{2}{c|}{\demph{random init.}}
& \resrand{39.4}{} & \resrand{78.5}{} & \resrand{35.1}{}  % coco pose
& \\
super. & IN-1M
& \ressup{48.3}{} & \ressup{85.6}{} & \ressup{50.6}{}  % coco pose
& \\
\hline
\textbf{MoCo} & IN-1M
& \reshl{50.1}{+}{1.8} & \reshl{86.8}{+}{1.2} & \reshl{53.9}{+}{3.3}  % coco pose
& \\
\textbf{MoCo} & IG-1B
& \reshl{50.6}{+}{2.3} & \reshl{87.0}{+}{1.4} & \reshl{54.3}{+}{3.7}  % coco pose
& \\
\end{tabular}
\end{raggedleft}
\vspace{.3em}

%% lvis
\begin{raggedleft}
\tablestyle{2pt}{1.1}
\begin{tabular}{ry{28}|x{48}x{48}x{48} c}
\multicolumn{2}{c|}{} &
\multicolumn{3}{c}{\fontsize{7.5pt}{1em}\selectfont \textbf{LVIS v0.5 instance segmentation}} & 
\\
\multicolumn{2}{c|}{pre-train} &
\apmask{~} & \apmask{50} & \apmask{75} &  % lvis 
\\
\shline
\multicolumn{2}{c|}{\demph{random init.}}
& \resrand{22.5}{} & \resrand{34.8}{} & \resrand{23.8}{}  % lvis mask
& \\
super. & IN-1M$^\dagger$
& \ressup{24.4}{} & \ressup{37.8}{} & \ressup{25.8}{}  % lvis mask
& \\
\hline
\textbf{MoCo} & IN-1M
& \res{24.1}{-}{0.3} & \res{37.4}{-}{0.4} & \res{25.5}{-}{0.3}  % lvis mask
& \\
\textbf{MoCo} & IG-1B
& \reshl{24.9}{+}{0.5} & \res{38.2}{+}{0.4} & \reshl{26.4}{+}{0.6}  % lvis mask
& \\
\end{tabular}
\end{raggedleft}
\vspace{.3em}

%% lvis
\begin{raggedleft}
\tablestyle{2pt}{1.1}
\begin{tabular}{ry{24}|x{40}x{40}|x{40}|x{36} c}
\multicolumn{2}{c|}{} &
\multicolumn{2}{c|}{\fontsize{7.5pt}{1em}\selectfont \textbf{Cityscapes instance seg.}} & 
\multicolumn{2}{c}{\fontsize{7.5pt}{1em}\selectfont \textbf{Semantic seg.} (mIoU)} & 
\\
\multicolumn{2}{c|}{pre-train} &
\apmask{~} & \apmask{50} &  % city mask 
{\fontsize{7.5pt}{1em}\selectfont \textbf{Cityscapes}} & 
{\fontsize{7.5pt}{1em}\selectfont \textbf{VOC}} &
\\
\shline
\multicolumn{2}{c|}{\demph{random init.}}
& \resrand{25.4}{} & \resrand{51.1}{}  % city mask
& \resrand{65.3}{}  % city seg
& \resrand{39.5}{}  % voc seg
& \\
super. & IN-1M
& \ressup{32.9}{} & \ressup{59.6}{}  % city mask
& \ressup{74.6}{}  % city seg
& \ressup{74.4}{}  % voc seg
& \\
\hline
\textbf{MoCo} & IN-1M
& \res{32.3}{-}{0.6} & \res{59.3}{-}{0.3}  % city mask
& \reshl{75.3}{+}{0.7}  % city seg
& \res{72.5}{-}{1.9}  % voc seg
& \\
\textbf{MoCo} & IG-1B
& \res{32.9}{{\transparent{0}+}}{0.0} & \reshl{60.3}{+}{0.7}  % city mask
& \reshl{75.5}{+}{0.9}  % city seg
& \res{73.6}{-}{0.8}  % voc seg
& \\
\end{tabular}
\end{raggedleft}
\vspace{.3em}
\caption{\textbf{MoCo \vs ImageNet supervised pre-training, fine-tuned on various tasks}.
For each task, the same architecture and schedule are used for all entries (see \appdx).
In the brackets are the gaps to the ImageNet supervised pre-training counterpart. In green are the gaps of at least {\fontsize{8pt}{1em}\selectfont \hl{${+}$\textbf{0.5}}} point.
\newline
{\footnotesize $^\dagger$: this entry is with BN frozen, which improves results; see main text.}}
\label{tab:more_tasks}
\vspace{-1.5em}
\end{table}
%##################################################################################################

\vspace{-.5em}
\subsubsection{More Downstream Tasks}

Table~\ref{tab:more_tasks} shows more downstream tasks (implementation details in \appdx). Overall, MoCo performs competitively with ImageNet supervised pre-training:

\emph{COCO keypoint detection}: supervised pre-training has \emph{no} clear advantage over random initialization, whereas MoCo outperforms in all metrics. 

\emph{COCO dense pose estimation} \cite{AlpGueler2018}: MoCo substantially outperforms supervised pre-training, \eg, by 3.7 points in \apdp{75}, in this highly localization-sensitive task.

\emph{LVIS v0.5 instance segmentation} \cite{Gupta2019}: this task has $\app$1000 long-tailed distributed categories.
Specifically in LVIS for the ImageNet supervised baseline, we find fine-tuning with frozen BN (24.4 \apmask{}) is better than tunable BN (details in \appdx). So we compare MoCo with the better supervised pre-training variant in this task. MoCo with IG-1B surpasses it in all metrics.

\mbox{\emph{Cityscapes instance segmentation} \cite{Cordts2016}: MoCo with IG-1B} is on par with its supervised pre-training counterpart in \apmask{}, and is higher in \apmask{50}.

\emph{Semantic segmentation}: On Cityscapes \cite{Cordts2016}, MoCo outperforms its supervised pre-training counterpart by up to 0.9 point. But on VOC semantic segmentation, MoCo is worse by at least 0.8 point, a negative case we have observed.

\paragraph{Summary.}
In sum, MoCo can \textbf{\emph{outperform}} its ImageNet supervised pre-training counterpart in \textbf{\emph{7}} detection or segmentation tasks.\footnote{Namely, object detection on VOC/COCO, instance segmentation on COCO/LVIS, keypoint detection on COCO, dense pose on COCO, and semantic segmentation on Cityscapes.}
Besides, MoCo is on par on Cityscapes instance segmentation, and lags behind on VOC semantic segmentation; we show another comparable case on iNaturalist \cite{VanHorn2018} in \appdx. Overall, \emph{MoCo has largely closed the gap between unsupervised and supervised representation learning in multiple vision tasks}.

Remarkably, in all these tasks, MoCo pre-trained on \mbox{IG-1B} is consistently better than MoCo pre-trained on \mbox{IN-1M}.
This shows that \emph{MoCo can perform well on this large-scale, relatively uncurated dataset}. This represents a scenario towards \emph{real-world} unsupervised learning.

\section{Discussion and Conclusion}

Our method has shown positive results of unsupervised learning in a variety of computer vision tasks and datasets. A few open questions are worth discussing.
MoCo's improvement from IN-1M to IG-1B is consistently noticeable but relatively small, suggesting that the larger-scale data may not be fully exploited. We hope an advanced pretext task will improve this.
Beyond the simple instance discrimination task \cite{Wu2018a}, it is possible to adopt MoCo for pretext tasks like masked auto-encoding, \eg, in language \cite{Devlin2019} and in vision \cite{Oord2018}. We hope MoCo will be useful with other pretext tasks that involve contrastive learning.

%##################################################################################################
%{\small
%\renewcommand\UrlFont{\color{Gray}\ttfamily}
%\bibliographystyle{ieee_fullname}
%\bibliography{moco}
%}
%\clearpage
%\end{document}
%##################################################################################################

\appendix
\section{Appendix}

% reset Figure number: e.g., Figure B.1.
\renewcommand\thefigure{\thesection.\arabic{figure}}
\renewcommand\thetable{\thesection.\arabic{table}}
\setcounter{figure}{0} 
\setcounter{table}{0} 

% reset table
\setcounter{table}{0}
\renewcommand{\thetable}{A.\arabic{table}}

\subsection{Implementation: Object detection backbones}
The R50-dilated-C5 and R50-C4 backbones are similar to those available in \texttt{Detectron2}~\cite{Wu2019}: (i) \emph{R50-dilated-C5}: the backbone includes the ResNet conv$_\text{5}$ stage with a dilation of 2 and stride 1, followed by a 3$\times$3 convolution (with BN) that reduces dimension to 512. The box prediction head consists of two hidden fully-connected layers. (ii) \emph{R50-C4}: the backbone ends with the conv$_\text{4}$ stage, and the box prediction head consists of the conv$_\text{5}$ stage (including global pooling) followed by a BN layer.

\subsection{Implementation: COCO keypoint detection}
We use Mask R-CNN (keypoint version) with R50-FPN, implemented in \cite{Wu2019}, fine-tuned on COCO \texttt{train2017} and evaluated on \texttt{val2017}. The schedule is 2$\x$.

\subsection{Implementation: COCO dense pose estimation}
We use DensePose R-CNN \cite{AlpGueler2018} with R50-FPN, implemented in \cite{Wu2019}, fine-tuned on COCO \texttt{train2017} and evaluated on \texttt{val2017}.
The schedule is ``s1$\x$''.

\subsection{Implementation: LVIS instance segmentation}

We use Mask R-CNN with R50-FPN, fine-tuned in LVIS \cite{Gupta2019} \texttt{train\_v0.5} and evaluated in \texttt{val\_v0.5}. We follow the baseline in \cite{Gupta2019} (arXiv v3 Appendix B).

LVIS is a new dataset and model designs on it are to be explored. The following table includes the relevant ablations (all are averages of 5 trials):
%##################################################################################################
\begin{center}
\vspace{-0.3em}
\small
% ------------------------------------------------
\tablestyle{5pt}{1.1}
\begin{tabular}{x{48}c|x{14}x{14}x{14}|x{14}x{14}x{14}}
& & \multicolumn{3}{c|}{1$\x$ schedule} & \multicolumn{3}{c}{2$\x$ schedule} \\
pre-train & BN &
\apmask{~} &
\apmask{50} &
\apmask{75} &
\apmask{~} &
\apmask{50} &
\apmask{75} \\
\shline
\supimgnet & frozen & 24.1 & 37.3 & 25.4 & 24.4 & 37.8 & 25.8 \\
\supimgnet & tuned & 23.5 & 36.6 & 24.8 & 23.2 & 36.0 & 24.4 \\
\hline
\mocoimgnet & tuned & 23.2 & 36.0 & 24.7 & 24.1 & 37.4 & 25.5 \\
\mocoins & tuned & 24.3 & 37.4 & 25.9 & \textbf{24.9} & \textbf{38.2} & \textbf{26.4} \\
\end{tabular}	
\vspace{-0.3em}
\end{center}
%##################################################################################################
A supervised pre-training baseline, end-to-end tuned but with BN frozen, has 24.4 \apmask{}. But tuning BN in this baseline leads to worse results and overfitting (this is unlike on COCO/VOC where tuning BN gives better or comparable accuracy).
MoCo has 24.1 \apmask{} with IN-1M and 24.9 \apmask{} with IG-1B, both outperforming the supervised pre-training counterpart under the same tunable BN setting. Under the best individual settings, MoCo can still outperform the supervised pre-training case (24.9 \vs 24.4, as reported in Table~\ref{tab:more_tasks} in Sec~\ref{sec:transfer}).

\subsection{Implementation: Semantic segmentation}

We use an FCN-based \cite{Long2015} structure. The backbone consists of the convolutional layers in R50, and the 3$\x$3 convolutions in conv$_5$ blocks have dilation 2 and stride 1. This is followed by two extra 3$\x$3 convolutions of 256 channels, with BN and ReLU, and then a 1$\x$1 convolution for per-pixel classification. The total stride is 16 (FCN-16s \cite{Long2015}). We set \mbox{dilation $=$ 6} in the two extra 3$\x$3 convolutions, following the large field-of-view design in \cite{Chen2017}.

Training is with random scaling (by a ratio in [0.5, 2.0]), cropping, and horizontal flipping.
The crop size is 513 on VOC and 769 on Cityscapes \cite{Chen2017}.
Inference is performed on the original image size. We train with mini-batch size 16 and weight decay 0.0001. Learning rate is 0.003 on VOC and is 0.01 on Cityscapes (multiplied by 0.1 at 70-th and 90-th percentile of training).
For VOC, we train on the \texttt{train\_aug2012} set (augmented by \cite{Hariharan2011}, 10582 images) for 30k iterations, and evaluate on \texttt{val2012}. For Cityscapes, we train on the \texttt{train\_fine} set (2975 images) for 90k iterations, and evaluate on the \texttt{val} set. Results are reported as averages over 5 trials.

\subsection{iNaturalist fine-grained classification}
In addition to the detection/segmentation experiments in the main paper, we study fine-grained classification on the iNaturalist 2018 dataset \cite{VanHorn2018}. We fine-tune the pre-trained models end-to-end on the \texttt{train} set ($\app$437k images, 8142 classes) and evaluate on the \texttt{val} set. Training follows the typical ResNet implementation in PyTorch with 100 epochs. Fine-tuning has a learning rate of 0.025 (\vs 0.1 from scratch) decreased by 10 at the 70-th and 90-th percentile of training.
The following is the R50 result:
\begin{center}
\vspace{-.3em}
\tablestyle{1pt}{1.2}	
\begin{tabular}{x{46}|x{45}x{45}x{45}x{45}}
pre-train & rand init. & super.$_\text{IN-1M}$ & \textbf{MoCo}$_\text{IN-1M}$ & \textbf{MoCo}$_\text{IG-1B}$  \\
\shline
accuracy (\%) & 61.8 & \textbf{66.1} & 65.6 & 65.8
\end{tabular}
\vspace{-.5em}
\end{center}
MoCo is $\app$4\% better than training from random initialization, and is closely comparable with its ImageNet supervised counterpart. This again shows that MoCo unsupervised pre-training is competitive.

\subsection{Fine-tuning in ImageNet}
Linear classification on frozen features (Sec.~\ref{sec:imagenet}) is a common protocol of evaluating unsupervised pre-training methods. However, in practice, it is more common to fine-tune the features end-to-end in a downstream task. 
For completeness, the following table reports end-to-end fine-tuning results for the 1000-class ImageNet classification, compared with training from scratch (fine-tuning uses an initial learning rate of 0.03, \vs 0.1 from scratch):
\begin{center}
\vspace{-.3em}
\tablestyle{4pt}{1.2}	
\begin{tabular}{x{56}|x{48}x{48}}
pre-train & random init. & \textbf{MoCo}$_\text{IG-1B}$ \\
\shline
accuracy (\%) & 76.5 & \textbf{77.3}
\end{tabular}
\vspace{-.3em}
\end{center}
As here ImageNet is the downstream task, the case of MoCo pre-trained on IN-1M does not represent a real scenario (for reference, we report that its accuracy is 77.0\% after fine-tuning).
But unsupervised pre-training in the \emph{separate}, \mbox{unlabeled} dataset of IG-1B represents a typical scenario: in this case, MoCo improves by 0.8\%. 

%##################################################################################################
%\newcommand{\cres}[1]{#1}
%\newcommand{\cresrand}[1]{#1}

%\newcommand{\cgap}[2]{
%\fontsize{6pt}{1em}\selectfont{(${#1}${#2})}
%}
%\newcommand{\cgaphl}[2]{
%\fontsize{6pt}{1em}\selectfont{\textcolor{Highlight}{(${#1}$\textbf{#2})}}
%}

\renewcommand{\randinit}{\tablestyle{1pt}{1} \begin{tabular}{z{21}y{26}} \multicolumn{2}{c}{random init.} \end{tabular}}

\begin{table*}[t]
\vspace{-1.em}
\small
% ------------------------------------------------
\hspace{-1.5em}
\resizebox{1.05\linewidth}{!}{
\subfloat[Mask R-CNN, R50-FPN, \textbf{2$\x$} schedule]{
\tablestyle{.8pt}{1.05}
\begin{tabular}{cr|
z{17}y{18}
z{17}y{18}
z{17}y{18}|
z{17}y{18}
z{17}y{18}
z{17}y{18}c
}
pre-train & ~ &
\multicolumn{2}{c}{\apbbox{~}} &
\multicolumn{2}{c}{\apbbox{50}} &
\multicolumn{2}{c|}{\apbbox{75}} &
\multicolumn{2}{c}{\apmask{~}} &
\multicolumn{2}{c}{\apmask{50}} &
\multicolumn{2}{c}{\apmask{75}} &\\
\shline
\demph{\randinit} & ~ &
\demph{36.7} & ~  & \demph{56.7} & ~  & \demph{40.0} & ~  &
\demph{33.7} & ~  & \demph{53.8} & ~  & \demph{35.9} & ~  & \\
\supimgnet & ~ &
40.6 & ~ & 61.3 & ~ & 44.4 & ~ &
36.8 & ~ & 58.1 & ~ & 39.5 & ~ & \\
\hline
\mocoimgnet & ~ &
40.8 & \cgap{+}{0.2} & 61.6 & \cgap{+}{0.3} & 44.7 & \cgap{+}{0.3} &
36.9 & \cgap{+}{0.1} & 58.4 & \cgap{+}{0.3} & 39.7 & \cgap{+}{0.2} & \\
\mocoins & ~ &
41.1 & \cgaphl{+}{0.5} & 61.8 & \cgaphl{+}{0.5} & 45.1 & \cgaphl{+}{0.7} & 
37.4 & \cgaphl{+}{0.6} & 59.1 & \cgaphl{+}{1.0} & 40.2 & \cgaphl{+}{0.7} & \\
\end{tabular}	
}  % end of subfloat
% ------------------------------------------------

% ------------------------------------------------
\subfloat[Mask R-CNN, R50-FPN, \textbf{6$\x$} schedule]{
\tablestyle{.8pt}{1.05}
\begin{tabular}{
z{17}y{18}
z{17}y{18}
z{17}y{18}|
z{17}y{18}
z{17}y{18}
z{17}y{18}c
}
%	pre-train &
\multicolumn{2}{c}{\apbbox{~}} &
\multicolumn{2}{c}{\apbbox{50}} &
\multicolumn{2}{c|}{\apbbox{75}} &
\multicolumn{2}{c}{\apmask{~}} &
\multicolumn{2}{c}{\apmask{50}} &
\multicolumn{2}{c}{\apmask{75}} &\\
\shline
%\demph{\randinit}
\demph{41.4} & ~  & \demph{61.9} & ~  & \demph{45.1} & ~  &
\demph{37.6} & ~  & \demph{59.1} & ~  & \demph{40.3} & ~  & \\
%\supimgnet &
41.9 & ~ & 62.5 & ~ & 45.6 & ~ &
38.0 & ~ & 59.6 & ~ & 40.8 & ~ & \\
\hline
%\mocoimgnet &
42.3 & \cgap{+}{0.4} & 62.7 & \cgap{+}{0.2} & 46.2 & \cgaphl{+}{0.6} &
38.3 & \cgap{+}{0.3} & 60.1 & \cgaphl{+}{0.5} & 41.2 & \cgap{+}{0.4} & \\
%\mocoins &
42.8 & \cgaphl{+}{0.9} & 63.2 & \cgaphl{+}{0.7} & 47.0 & \cgaphl{+}{1.4} & 
38.7 & \cgaphl{+}{0.7} & 60.5 & \cgaphl{+}{0.9} & 41.3 & \cgaphl{+}{0.5} & \\
\end{tabular}	
}  % end of subfloat
% ------------------------------------------------
}  % end of resizebox

% ------------------------------------------------
\vspace{.3em}
\caption{Object detection and instance segmentation fine-tuned on COCO: \textbf{2$\times$ \vs 6$\times$ schedule}. In the brackets are the gaps to the ImageNet supervised pre-training counterpart. In green are the gaps of at least {\fontsize{8pt}{1em}\selectfont \hl{${+}$\textbf{0.5}}} point.
}
\label{tab:coco_longer}
\vspace{-1.em}
\end{table*}
%##################################################################################################

\subsection{COCO longer fine-tuning}

In Table~\ref{tab:coco} we reported results of the 1$\times$ ($\app$12 epochs) and 2$\times$ schedules on COCO. These schedules were inherited from the original Mask R-CNN paper \cite{He2017}, which could be suboptimal given later advance in the field. In Table~\ref{tab:coco_longer}, we supplement the results of a 6$\times$ schedule ($\app$72 epochs) \cite{He2019} and compare with those of the 2$\times$ schedule.

We observe: (i) fine-tuning with ImageNet-supervised pre-training still has improvements (41.9 \apbbox{}); (ii) training from scratch largely catches up (41.4 \apbbox{}); (iii) the MoCo counterparts improve further (\eg, \mbox{to 42.8~\apbbox{}}) and have larger gaps (\eg, +0.9~\apbbox{} with 6$\times$, \vs +0.5~\apbbox{} with 2$\times$). 
Table~\ref{tab:coco_longer} and Table~\ref{tab:coco} suggest that the MoCo pre-trained features can have \emph{larger} advantages than the ImageNet-supervised features when fine-tuning \emph{longer}.

\subsection{Ablation on Shuffling BN}

\begin{figure}[h]
\centering
\includegraphics[width=0.98\linewidth]{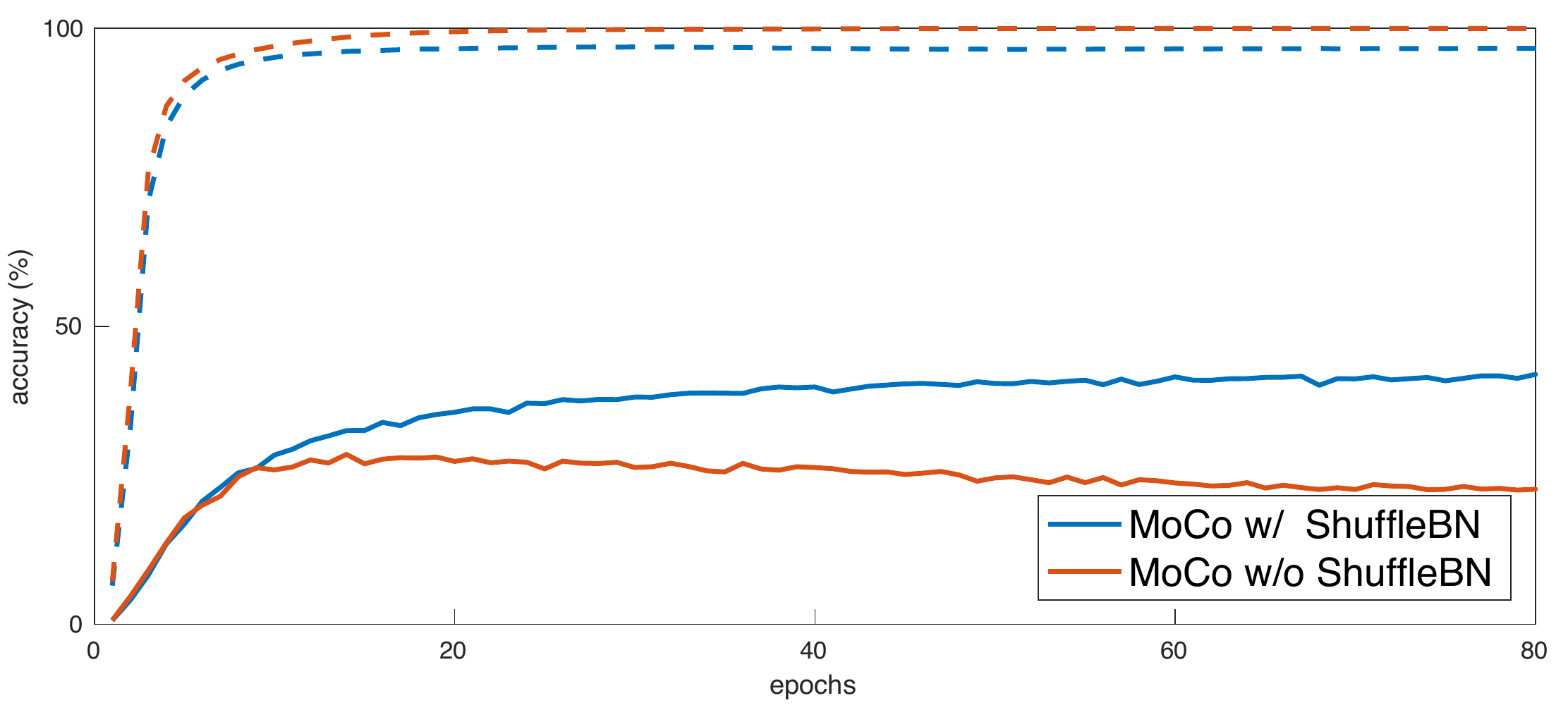}
\caption{\textbf{Ablation of Shuffling BN}. \emph{Dash}: training curve of the pretext task, plotted as the accuracy of ($K$+1)-way dictionary lookup. \emph{Solid}: validation curve of a kNN-based monitor \cite{Wu2018a} (not a linear classifier) on ImageNet classification accuracy. This plot shows the first 80 epochs of training: training longer without shuffling BN overfits more.}
\label{fig:shufflebn}
\end{figure}

Figure~\ref{fig:shufflebn} provides the training curves of MoCo with or without shuffling BN:
removing shuffling BN shows obvious overfitting to the pretext task: training accuracy of the pretext task (dash curve) quickly increases to $>$99.9\%, and the kNN-based validation classification accuracy (solid curve) drops soon.
This is observed for both the MoCo and end-to-end variants; the memory bank variant implicitly has different statistics for $q$ and $k$, so avoids this issue.

These experiments suggest that without shuffling BN, the sub-batch statistics can serve as a ``signature" to tell which sub-batch the positive key is in. Shuffling BN can remove this signature and avoid such cheating.

% # ------------------------------
% # ------------------------------

{\small
\renewcommand\UrlFont{\color{Gray}\ttfamily}
\bibliographystyle{ieee_fullname}
\bibliography{moco}
}

\end{document}